%% file: main.tex
\documentclass[11pt,letterpaper]{article}
\usepackage[hyperref]{acl2018}
\usepackage{times}
\usepackage{latexsym}
\usepackage{amsmath}
\usepackage{amssymb}
\usepackage{relsize}
\usepackage{xcolor}
\usepackage{xspace}
\usepackage{epsfig,url,algorithm,algorithmic,multirow}

\usepackage{multirow}
\usepackage{bigstrut}
\usepackage{color, colortbl}
\definecolor{LightCyan}{rgb}{0.88,1,1}
\newcolumntype{g}{>{\columncolor{LightCyan}}c}

\definecolor{Orange}{rgb}{1,0.7,0}
\newcolumntype{o}{>{\columncolor{Orange}}c}

\definecolor{Blou}{rgb}{0,0.7,1}
\newcolumntype{b}{>{\columncolor{Blou}}c}

\aclfinalcopy 

\newcommand{\mat}[1]{\boldsymbol{\mathbf{#1}}}

\newcommand{\squishlist}{
	\begin{list}{$\bullet$}
		{ \setlength{\itemsep}{0pt}
			\setlength{\parsep}{3pt}
			\setlength{\topsep}{3pt}
			\setlength{\partopsep}{0pt}
			\setlength{\leftmargin}{1.5em}
			\setlength{\labelwidth}{1em}
			\setlength{\labelsep}{0.5em} } }
		
\newcommand{\squishend}{\end{list}}

\DeclareMathOperator*{\argmin}{arg\,min}
\DeclareMathOperator*{\argmax}{arg\,max}

\title{Characterizing Departures from Linearity in  Word Translation}

\author{Ndapa Nakashole \and Raphael Flauger \\
	Computer Science and Engineering\\
	University of California, San Diego\\
	La Jolla, CA 92093\\
	{\tt nnakashole@eng.ucsd.edu}}

\date{}

\begin{document}

\maketitle

\begin{abstract}
We investigate the behavior of  maps learned by  machine translation methods. The maps translate words by projecting between word embedding spaces of  different languages. 
We locally approximate these maps using linear maps, and find that they vary across the word embedding space. This demonstrates that the underlying maps are non-linear. Importantly, we show that the locally linear maps vary by an amount that is tightly correlated with the distance between the neighborhoods on which they are trained.
Our results can be used to test non-linear methods, and to drive  the design of more accurate maps for word translation.


\end{abstract}

\input{introduction.tex}
\input{relatedwork.tex}
\input{global.tex}

\input{experiments.tex}

\section{Conclusions}
In this paper, we provide evidence that the  assumption of linearity made by a large body of  current work on cross-lingual mapping for word translation does not hold. We locally approximate the underlying non-linear map using linear maps, and show that these maps  vary across neighborhoods in vector  space by an amount that is tightly correlated with the distance between the neighborhoods on which they are trained. These results can be used to test non-linear methods. We leave using the findings of this paper to design more accurate maps  as future work. 


\section*{Acknowledgments}
We thank the anonymous reviewers for their constructive
comments. We also gratefully acknowledge Amazon for AWS Cloud Credits for Research, and Nvidia for a GPU grant.

\clearpage

\nocite{*} 
\bibliography{linearity}
\bibliographystyle{acl_natbib}

\end{document}

%% file: introduction.tex
\section{Introduction}
 
Following the success of monolingual word embeddings  \cite{collobert2011natural}, 
a number of studies have recently explored multilingual word embeddings. The goal is to  learn word vectors such that similar words have similar vector representations  regardless of their language \cite{zou2013bilingual,DBLP:conf/acl/UpadhyayFDR16}. Multilingual word embeddings have applications in  machine translation, and hold promise for cross-lingual model transfer in NLP tasks such as parsing or part-of-speech tagging.

A class of methods has emerged  whose core technique  is to learn  \textit{linear} maps between  vector spaces of different languages \cite{DBLP:journals/corr/MikolovLS13,DBLP:conf/eacl/FaruquiD14,vulic2016role, artetxe2016learning,conneau2017word}. These methods work as follows: For a given pair of languages, first, monolingual word vectors are learned
independently for each language, and second, under the assumption that  word vector  spaces exhibit comparable structure across languages, a linear mapping function is learned 
to connect the two monolingual vector spaces. The map can then be used to translate words between the  language pair.

\begin{figure}[t]
	\begin{center}
		\includegraphics{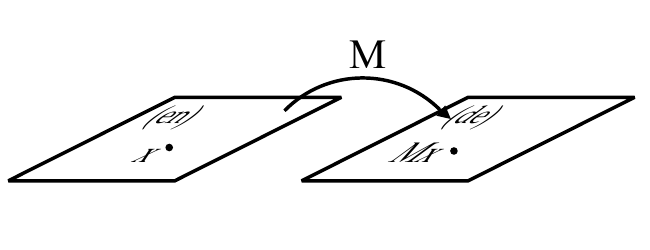}
		\includegraphics{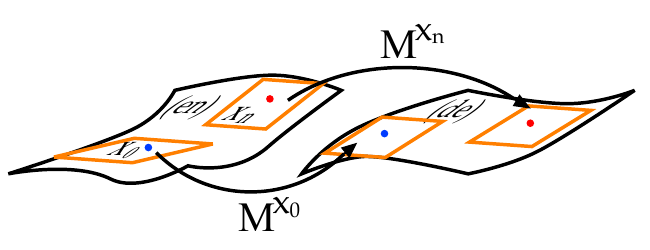}
		\caption{\textbf{Top:} Assumption of linearity implies a single linear map $\mat{M}$. 
		\textbf{Bottom:} Our hypothesis is that the underlying map  is expected to be non-linear but in small enough neighborhoods can be approximated by linear maps $\mat{M^{x_i}}$ for each neighborhood defined by $\mat{x_i}$.}
		\label{fig:embeddingspaces}
	\end{center}
\end{figure}

Both seminal \cite{DBLP:journals/corr/MikolovLS13}, and   state-of-the-art methods \cite{conneau2017word}  found  linear maps to substantially outperform specific non-linear maps generated by feedforward neural networks.
Advantages of linear maps include:
\textit{ 1)} In settings with limited  training data, accurate linear maps can still be learned \cite{conneau2017word,zhang2017adversarial,artetxe2017learning,smith2017offline}. For example, in unsupervised learning, \cite{conneau2017word}  found that using non-linear mapping functions  made adversarial training  unstable\footnote{https://openreview.net/forum?id=H196sainb}.
\textit{ 2)}  One can  easily impose constraints  on the linear maps at training time  to ensure that the quality of the  monolingual embeddings is preserved after mapping \cite{xing2015normalized,smith2017offline}. 

However, it is not well understood to what extent the assumption of linearity holds and how it affects performance. In this paper, we investigate the behavior of word translation maps, and show that there is clear evidence of  departure from linearity. 

Non-linear maps beyond those generated by feedforward neural networks have also been explored for this task \cite{lu2015deep,shi2015learning,wijaya2017learning,shi2015learning}. However, no attempt was made to characterize the resulting maps. 

In this paper, we allow for an underlying mapping function that is non-linear, but assume that it can be approximated by linear maps at least in small enough neighborhoods. If the underlying map is  linear,  all local approximations should be identical, or, given the finite size of the training data, similar. In contrast, if the underlying map is non-linear,  the locally linear approximations will depend on the neighborhood. Figure~\ref{fig:embeddingspaces}  illustrates the difference between the assumption of a single linear map, and our working hypothesis of locally linear approximations to a non-linear map. The variation of the linear approximations provides a characterization of the nonlinear map. We show that the local linear approximations vary across neighborhoods in the embedding space by an amount that is tightly correlated with the distance between the neighborhoods on which they are trained. The functional form of this variation can be used to test non-linear methods.

%% file: relatedwork.tex
\section{Review of Prior Work}\label{preliminaries}

To learn  linear word translation maps, different  loss functions have been proposed.
The simplest is the regularized least squares loss, where the linear map  ${\mat{M}}$ is learned as follows: 
$\hat{\mat{M}}~=~\argmin_{\mat{M} }||\mat{M}\mat{X}~-~\mat{Y}||_F~+~\lambda||\mat{M}||$,
here $\mat{X}$ and $\mat{Y}$   are  matrices that contain word embedding vectors for the source    and   target   language \cite{DBLP:journals/corr/MikolovLS13,dinu2014improving,vulic2016role}.
The translation $t$ of a source language word $s$ is then given by: $t~=~\argmax_{t} \cos(\mat{M}x_s, y_t)$.

\cite{xing2015normalized}  obtained improved results by imposing an orthogonality constraint on $\mat{M}$,  by minimizing
$||\mat{M}\mat{M}^T~- ~\mat{I}||$
where $\mat{I}$ is the identify matrix. 
Another loss function used in prior work is the max-margin  loss,
which has  been shown to significantly outperform the least squares loss   \cite{DBLP:conf/acl/LazaridouDB15,nakashole2017knowledge}. 

Unsupervised or limited supervision methods for learning word translation maps have recently been proposed  \cite{Barone2016,conneau2017word,zhang2017adversarial,artetxe2017learning,smith2017offline}. However, the underlying methods for learning the mapping function are similar to prior  work \cite{xing2015normalized}.

Non-linear cross-lingual  mapping methods have been proposed.  In \cite{wijaya2017learning}  when dealing with rare words, the proposed method backs-off to a feed-forward neural network. \cite{shi2015learning} model relations across languages.  \cite{lu2015deep} proposed a deep canonical correlation analysis based  mapping method.  Work  on phrase translation has explored the use of many local maps  that are individually trained  \cite{zhao2015learning}. 
In contrast to our work, these prior  papers do not attempt to characterize the behavior of the  resulting maps.


Our hypothesis 
is similar in spirit  to the use of  locally linear embeddings
for nonlinear dimensionality
reduction \cite{roweis2000nonlinear}.

%% file: global.tex
\section{Neighborhoods in  Word Vector Space}
In order to  study  the behavior of  word translation maps, we begin by  introducing a simple notion of neighborhoods in the embedding space.
For a  given language (e.g., English, $en$), we  define a  \textit{neighborhood} of a word  as follows: First, we pick a word~$x_i$,  whose corresponding vector is $x_i ~\in ~\mat{X}^{en}$, as an anchor. Second,  we  initialize  a neighborhood  $\mathcal{N}(x_i)$ containing a single vector $x_i$. We then   grow the neighborhood by adding all words whose cosine similarity  to  $x_i$ is  $\geq s$. The resulting neighborhood is defined as:
$\mathcal{N}(x_i, s) = \{\, x_j \mid \cos(x_i, x_j)  \geq  s \,\}$.

Suppose we pick the word {\it multivitamins} as the anchor word.  We can generate  neighborhoods using $
\mathcal{N}(x_\textit{ multivitamins}, s)$  where for each value of $s$ we get a different neighborhood. Neighborhoods corresponding to larger values of $s$ are subsumed by those corresponding to smaller values of $s$.

\begin{figure}[t]
	\begin{center}
		\includegraphics[width=0.7\linewidth]{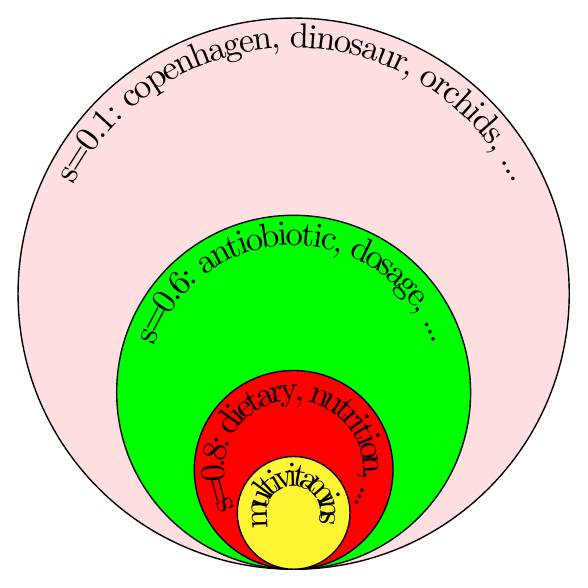}
		\vspace{-0.3cm}
		\caption{Neighborhoods formed around the word {``multivitamins"}.} 
		\label{fig:neighborhood_plot}
	\end{center}
\end{figure}

Figure \ref{fig:neighborhood_plot} illustrates the process of generating neighborhoods around the word {\it multivitamins}.
For  large values of $s$ (e.g., $s=0.8$),  the resulting neighborhoods only contain  words that are closely related to  the anchor word, such as {\it dietary} and {\it nutrition}. As $s$ gets smaller (e.g., $s=0.6$), the neighborhood gets larger, and  includes  words  that are less related to the anchor,  such as  {\it antibiotic}.

Using this simple  method, we can define different-sized neighborhoods around any word in the vocabulary.


\begin{table*}[t]
	\begin{center}
		\resizebox{\textwidth}{!}{%
			\begin{tabular}{|l|c|c|c|c| c| c| c| c| c|}
				\hline
				0 & 1 & 2 &3&4& 5& 6& 7 & 8 & 9\bigstrut \\
				\hline
				\hline
				\multirow{2}{0cm}{Anchor Word}& \multicolumn{2}{c|}{\centering Data, $\mathcal{N}(x_i, s=0.5)$} & $x_0$  Similarity  &  \multicolumn{4}{c|}{\centering Translation Accuracy}  & \multicolumn{2}{c|}{\centering Matrix Property}\bigstrut \\
				\cline{2-9} & \multicolumn{1}{c|}{Train} & \multicolumn{1}{c|}{Test} &  \multicolumn{1}{c|}{$\cos(x_0, x_i)$} &
				\multicolumn{1}{c|}{$\mat{M}$}  &
				\multicolumn{1}{c|}{$\mat{M^{x_0}}$}  & \multicolumn{1}{c|}{$\mat{M^{x_i}}$} &
				\multicolumn{1}{|g|}{$\Delta$}  &
				\multicolumn{1}{|g|}{ $\cos(\mat{M^{x_0}}, \mat{M^{x_i}})$} &
				\multicolumn{1}{|g|}{$||M||$}  \bigstrut \\ \hline
				$x_0$:multivitamins &3,415 & 500 & 1.0 & 58.3 & \textbf{68.2 } &\textbf{68.2}& 0 &1.0 & 33.07\bigstrut \\
				$x_1$:antibiotic &3,507 &500 &0.60 & 61.1 & 67.3  & \textbf{72.7}& $5.4\uparrow$ & 0.59 & 33.29\bigstrut \\
				$x_2$:disease &2,478 &500 &0.45 & 69.3 & 59.2 &\textbf{73.4}&$14.2 \uparrow$  &0.31 & 35.35 \bigstrut \\
				$x_3$:blowflies &2,434&500&0.33 & 71.4 & 28.4 &\textbf{73.2}&$44.8 \uparrow$ &0.20 &33.36\bigstrut \\
				$x_4$:dinosaur&990&500 &0.24 & 63.2 & 14.7 &\textbf{77.1}&$62.4 \uparrow$ &0.14 &36.50 \bigstrut \\
				$x_5$:orchids &2,981&500 &0.19& 73.7 &  19.3 &\textbf{78.0}&$58.7 \uparrow$ &0.20 & 30.68\bigstrut \\
				$x_6$:copenhagen &2,083 &500&0.11 & 38.5 & 31.2 &\textbf{67.4}&$36.2 \uparrow$  &0.15&31.42 \bigstrut \\
				
				\hline
			\end{tabular}
		}
	\end{center}
	\caption{The behavior of word translation maps trained on different neighborhoods ( en $\rightarrow$ de translation). Highlighted columns illustrate variations in maps.
		Accuracy refers to precision at 10.}
	\label{tbl:glinearexperiment}
\end{table*}

%% file: experiments.tex
\section{Analysis of Map Behavior}

%
%
Given the above  definition of neighborhoods, we  now  seek to  understand how word translation maps change as we move across neighborhoods in  word embedding space.

\paragraph{Questions Studied.} We study the following questions:
[Q.1] Is there a single linear map for word translation that produces the same level of  performance regardless of where in the vector space the words being translated fall? 
[Q.2] If there is no such single linear  map,  but instead multiple neighborhood-specific ones, is there a relationship between neighborhood-specific maps  and the distances between their respective neighborhoods?

\subsection{Experimental Setup and Data} 
In our first experiment we   translate  from English ($en$)   to   German ($de$). We obtained pre-trained word embeddings  from FastText \cite{bojanowski2017enriching}.  In the first experiment, we follow common practice \cite{DBLP:journals/corr/MikolovLS13, DBLP:journals/corr/AmmarMTLDS16,nakashole2017knowledge,vulic2016role}, and used the Google Translate API to obtain training and test data. We  make our  data available for reproducibility\footnote{\url{nakashole.com/mt.html}}.
For the  second  and third experiments, we  repeat the first experiment, but instead of  using Google Translate, we use the recently released Facebook AI Research dictionaries \footnote{\url{https://github.com/facebookresearch/MUSE}} for train  and test data.  The last two experiments were performed on a different language pairs:  English ($en$) to Portuguese ($pt$), English ($en$) to Swedish ($sv$).

In our all experiments, the cross-lingual maps are learned using the max-margin  loss,
which has  been shown to perform competitively, while having fast run-times.   \cite{DBLP:conf/acl/LazaridouDB15,nakashole2017knowledge}.
The max-margin loss  aims to rank  correct training data pairs  $(x_i,y_i)$  higher than incorrect pairs  $(x_i,y_j)$ with a margin of at least $\gamma$. The margin $\gamma$ is a hyper-parameter and  the incorrect labels,  $y_j$  can be selected randomly such that $j \neq i$ or  in a more  application specific manner. In our experiments, we set $\gamma = 0.4$ and randomly selected negative examples, one negative example for each training data point.

Given a seed dictionary as  training data of the form $D^{tr}=\{x_{i}, y_{i}\}^{m}_{i=1}$, the mapping function is 
\begin{eqnarray}
&\hskip -1cm\hat{\mat{W}} = \argmin\limits_{\mat{W} } \sum\limits_{i=1}^m  \sum\limits_{j\neq i}^k  \max\Big(0, \gamma  \nonumber \\
&\hskip 0.5cm + d(y_i,\mat{W}x_i)  - d(y_j,\mat{W}x_i)\Big),
\label{eqn:hingelossnew}
\end{eqnarray}

where $\hat{y_i}= \mat{W}x_i$ is the prediction, $k$ is the number of incorrect examples per training instance, and  $d(x,y)= (x-y)^{2}$ is the distance measure.


For the first experiment, we picked the following  words as anchor words and obtained  maps associated with each of their neighborhoods: $\mat{M}^\textit{ (multivitamins)}$,  $\mat{M}^\textit{ (antibiotic)}$, $\mat{M}^\textit{ (disease)}$, $\mat{M}^\textit{ (blowflies)}$, $\mat{M}^\textit{ (dinosaur)}$, $\mat{M}^\textit{ (orchids)}$, $\mat{M}^\textit{ (copenhagen)}$.  For each anchor word, we set $s=0.5$, thus the neighborhoods are  $\mathcal{N}(x_{i}, 0.5)$ where $x_i$ is the vector of the anchor word. 
The training data  for learning each neighborhood-specific linear map  consists of vectors  in  $\mathcal{N}(x_{i}, 0.5)$ and their translations.

Table~\ref{tbl:glinearexperiment}  shows details of the training  and test data  for each neighborhood.
The words shown in  Table~\ref{tbl:glinearexperiment} were picked as follows: first we picked the  word \textit{multivitamins}, then we  picked the other words to have varying degrees of similarity to it. The cosine similarity of these words to the word `multivitamins' are shown in column 3 of Table~\ref{tbl:glinearexperiment}. It is also worth noting that there is nothing special about these words. In fact, the last two  experiments were carried out on  different set of words, and on  different language pairs.

\begin{table*}[t]
	\begin{center}
			\begin{tabular}{|l|c|c|c|c| c| c|}
				\hline
				0 & 1 & 2 &3&4& 5& 6\bigstrut \\
				\hline
				\hline
				\multirow{2}{0cm}{Anchor Word}& \multicolumn{2}{c|}{\centering Data, $\mathcal{N}(x_i, s=0.5)$} & $x_0$  Similarity  &  \multicolumn{3}{c|}{\centering Translation Accuracy} \bigstrut \\
				\cline{2-7} & \multicolumn{1}{c|}{Train} & \multicolumn{1}{c|}{Test} &  \multicolumn{1}{c|}{$\cos(x_0, x_i)$} &
				\multicolumn{1}{c|}{$\mat{M}$}  &
				\multicolumn{1}{c|}{$\mat{M^{x_0}}$}  & \multicolumn{1}{c|}{$\mat{M^{x_i}}$} 
				\bigstrut \\ \hline
				$x_0$:clotting &1,908 & 300 &  1.0 & 80.7&\textbf{85.2} & \textbf{85.2} \bigstrut \\	
				$x_1$:heparin &1,487 & 300  & 0.79 & 74.3 &75.1& \textbf{75.4 }   \bigstrut \\
				$x_2$:inflammation &1,670 &300  &  0.75 & 85.3 &85.4 &\textbf{ 87.3 }  \bigstrut \\
				$x_3$:metabolites &1,590 & 300  & 0.66 & 77.3  &80.0 &\textbf{ 82.1 }   \bigstrut \\	
				$x_4$:hydroxides &1,279 & 300  &  0.54 & \textbf{77.3}  &67.3& {77.1}  \bigstrut \\
				$x_5$:giovannini &1,986 & 300  &  0.13 & 44.0  &11.7& \textbf{72.3}  \bigstrut \\
				$x_6$:gerardo &1,986 & 300  &  0.11 & 32.3  &3.6& \textbf{56.3}\bigstrut \\		
				
				\hline
			\end{tabular}
	\end{center}
	\caption{Different language pair ( en $\rightarrow$ pt) English to Portuguese, and different sets of neighborhoods.  Train and test is from the  FAIR/MUSE  word translation lexicons. Accuracy refers to precision at 10}
	\label{tbl:mapspt}
\end{table*}

\begin{table*}[ht]
	\begin{center}
			\begin{tabular}{|l|c|c|c|c| c| c|}
				\hline
				0 & 1 & 2 &3&4& 5& 6\bigstrut \\
				\hline
				\hline
				\multirow{2}{0cm}{Anchor Word}& \multicolumn{2}{c|}{\centering Data, $\mathcal{N}(x_i, s=0.5)$} & $x_0$  Similarity  &  \multicolumn{3}{c|}{\centering Translation Accuracy} \bigstrut \\
				\cline{2-7} & \multicolumn{1}{c|}{Train} & \multicolumn{1}{c|}{Test} &  \multicolumn{1}{c|}{$\cos(x_0, x_i)$} &
				\multicolumn{1}{c|}{$\mat{M}$}  &
				\multicolumn{1}{c|}{$\mat{M^{x_0}}$}  & \multicolumn{1}{c|}{$\mat{M^{x_i}}$}  \bigstrut \\ \hline
				$x_0$:species &1,765 & 200  &  1.0 & 58.0 &\textbf{60.0} &\textbf{ 60.0 } \bigstrut \\
				$x_1$:genus &1,374 & 200  &  0.77 & \textbf{56.4} &53.1 &{56.0} \bigstrut \\
				$x_2$:laticeps & 1,868  & 200 & 0.60 & \textbf{81.0 } &73.4 & {79.1 }  \bigstrut \\	
				$x_3$:femoralis &1,689 & 200  &  0.52 & \textbf{85.3 }&76.3& {83.4 } \bigstrut \\
				$x_4$:coneflower &1,077 & 200  &  0.47 & 39.6  &31.6 & \textbf{43.0 } \bigstrut \\	
				$x_5$:epicauta &1,339 & 200 &  0.43 & 53.4  &42.1  & \textbf{54.2} \bigstrut \\	
				$x_6$:kristoffersen &1,227 & 200  &  0.09 & 24.3 &10.3 & \textbf{57.1 } \bigstrut \\	
				\hline	
			\end{tabular}
	\end{center}
	\caption{Different language pair ( en $\rightarrow$ sv) English to Swedish, and different sets of neighborhoods.  Train and test is from the  FAIR/MUSE  word translation lexicons. Accuracy refers to precision at 10.}
	\label{tbl:glinearexperimentOtherWords2}
\end{table*}

\subsection{Map Similarity Analysis} If indeed there exists a map that is the same linear map everywhere,  we expect the above neighborhood-specific maps to be similar.  Our analysis makes use of the following definition of matrix similarity:
\begin{equation}
\cos(\mat{M_1}, \mat{M_2}) =\frac{tr(\mat{M_1}^T \mat{M_2})}  {\sqrt{tr(\mat{M_1}^T \mat{M_1}) tr(\mat{M_2}^T \mat{M_2}) } }
\label{eqn:matrixcosine}
\end{equation}

Here $tr(\mat{M})$ denotes the trace of the matrix $\mat{M}$.
$tr(\mat{M_1}^T \mat{M_1})$ computes the Frobenius norm $||\mat{M_1}||^2$, and $tr(\mat{M_1}^T\mat{M_2} )$ is the  Frobenius inner product.  That is,  $\cos(\mat{M_1}, \mat{M_2})$ computes the cosine similarity between the vectorized versions of matrices $\mat{M_1}$ and $\mat{M_2}$.

\subsection{Experimental Results} 
The main results of our analysis are shown in Table \ref{tbl:glinearexperiment}. 

We now analyze the results of Table \ref{tbl:glinearexperiment} in detail. The \textit{0th} column contains the anchor word, $x_i$,  around which the neighborhood is formed. The \textit{1st}, and \textit{2nd} columns contain the size of the training and test  data  from $\mathcal{N}(x_i, s=0.5)$ where $x_i$ is the word vector for the anchor word.  

\begin{figure}[t]
	\includegraphics[width=1.0\linewidth]{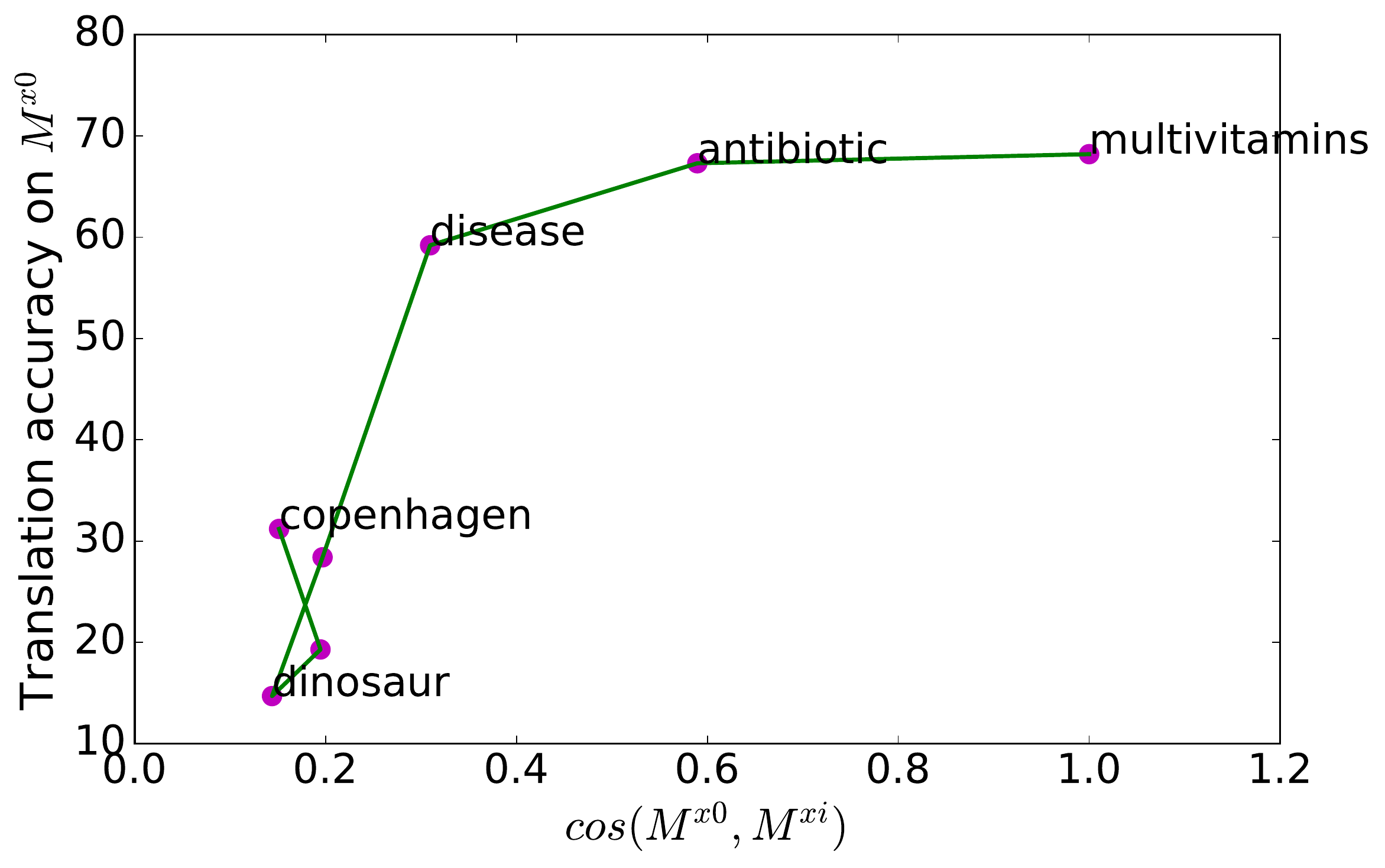}
	\vspace{-0.8cm}
	\caption{Correlation between   the \textit{8th} column (x-axis),  map similarity $\cos(\mat{M^{x_0}}, \mat{M^{x_i}})$, and  the \textit{5th} column (y-axis),  performance of map $M^{x_0}$ on test data from the neighborhood anchored at $x_i$.  }
	\label{fig:matrixsimilarity}
\end{figure}

The \textit{3rd} column contains the  cosine similarity between $x_0$, {\it multivitamins},  and $x_i$.  For example, $x_1$ ({\it antibiotic}) is the most similar to $x_0$  (0.6), and $x_6$, {\it copenhagen}, is the least similar to $x_0$ (0.11).

The \textit{4th} column is the translation accuracy of the single global map $M$, training on data from all  $x_i$ neighborhoods. The \textit{5th} column is the translation accuracy of the map $M^{x_0}$,  trained on the training data of $x_0$, and tested on the test data in $x_i$. We use precision at top-10 as a measure of translation accuracy. Going down this column we can see that accuracy is highest on the test data from the neighborhood anchored at $x_0$  itself, followed by neighborhoods  anchored  $x_1$ and $x_2$: antibiotic and disease.   Accuracy is lowest on the test data from the  neighborhoods  anchored  at words  further  away from $x_0$, in particular  $x_3$ to $x_6$: blowflies,  dinosaur, orchids,  copenhagen. 

The $6th$ column is translation accuracy  of  the  map $M^{x_i}$, trained on the training data of  the neighborhood anchored at $x_i$, and tested on the test data in $x_i$. We can see that compared to the $5th$ column, in all cases performance is higher  when we apply the map trained on  data from the neighborhood, $M^{x_i}$ instead of  $M^{x_0}$.
The $7th$ column shows the difference in translation accuracy of the map  $M^{x_i}$  and $M^{x_0}$.  This shows that  the  more dissimilar the neighborhood anchor word $x_i$ is from $x_0$ according to the cosine similarity shown in the $4rd$ column,  the larger this difference is. 

The local maps, $6th$ column,  $M^{x_i}$ in all cases outperform the global map $4th$ column, in Table~\ref{tbl:glinearexperiment}.

The  \textit{8th} column shows the  similarity between maps  $M^{x_i}$ and $M^{x_0}$ as computed by Equation \ref{eqn:matrixcosine}. This column shows that the similarity between these learned maps is highly correlated with the cosine similarity or distance between the words in $3rd$ column. We also see a correlation  with the translation accuracy in the $5th$ column. This correlation is visualized in Figure~\ref{fig:matrixsimilarity}.  Finally, the $9th$ column shows the magnitudes of the maps. The magnitudes vary somewhat between the  maps trained on the different neighborhoods, and are significantly different from the magnitude expected for an orthogonal matrix. (For an orthogonal $300\times300$ matrix $O$ the norm is $||O||=\sqrt{300}\approx 17$).

 In order to determine the generality of our results,  we carried out  the same experiment on different   language pairs and different sets of neighborhoods, as shown in Table \ref{tbl:mapspt} and Table \ref{tbl:glinearexperimentOtherWords2}. Crucially, we see the same trends  as those observed in Table \ref{tbl:glinearexperiment}. In particular, the key trends that maps vary by an amount that is tightly correlated with the distance between neighborhoods as reflected in5th and 6th  columns of Tables   \ref{tbl:mapspt} and \ref{tbl:glinearexperimentOtherWords2}. This   shows the generality of our findings in Table \ref{tbl:glinearexperiment}.
 
\paragraph{Experiments Summary.} In summary,  our experimental study suggests the following:
i)  linear maps vary across neighborhoods, implying that  the  assumption of a linear map does not  to hold. 
ii) the difference between maps is tightly correlated with the distance between neighborhoods.